\documentclass[12pt]{scrartcl}
\usepackage[T1]{fontenc}							
\usepackage[utf8]{inputenc}

\usepackage{geometry}
\usepackage{scrpage2}
	\automark[subsection]{subsection}
\usepackage{setspace}
	\spacing{1.2}
\usepackage{enumerate}
\usepackage{url}
\usepackage{array, multirow, dcolumn, booktabs}
\usepackage{rotating}

\usepackage{xcolor, graphicx}

\usepackage{amssymb}
\usepackage{amsmath, amsthm}

\KOMAoption{headings}{small}

\DeclareMathOperator{\WER}{WER}
\DeclareMathOperator{\CER}{CER}
\DeclareMathOperator*{\argmin}{arg min}
\DeclareMathOperator*{\argmax}{arg max}

\providecommand{\keywords}[1]{\textbf{\textit{Keywords --- }} #1}

\theoremstyle{definition}

\begin{document}

\title{CITlab ARGUS for historical handwritten documents}
\subtitle{Description of CITlab's System for the HTRtS 2014 Handwritten Text Recognition Task}
\author{Tobias Strauß \and Tobias Grüning \and Gundram Leifert \and Roger Labahn\thanks{corresponding author; CITlab, Institute of Mathematics, University of Rostock\newline\{tobias.gruening, gundram.leifert, tobias.strauss, roger.labahn\}@uni-rostock.de}}
\date{May 02, 2014}
\maketitle

\begin{abstract} 
We describe CITlab's recognition system for the HTRtS competition attached to the 14.\,International Conference on Frontiers in Handwriting Recognition, ICFHR 2014. The task comprises the recognition of historical handwritten documents.
The core algorithms of our system are based on multi-dimensional recurrent neural networks (MDRNN) and connectionist temporal classification (CTC).
The software modules behind that as well as the basic utility technologies are essentially powered by PLANET’s ARGUS framework for intelligent text recognition and image processing.
\end{abstract}
\keywords{MDRNN, LSTM, CTC, handwriting recognition, neural network}

\section{Introduction}
The International Conference on Frontiers in Handwriting Recognition ICFHR 2014\footnote{\url{http://www.icfhr2014.org}} hosts a variety of competitions in that area. Among others, the Handwritten Text Recognition on tranScriptorium Datasets (HTRtS) competition attracted our attention because we expected CITlab's handwriting recognition software to be able to successfully deal with the respective task.

HTRtS\footnote{\url{http://transcriptorium.eu/~htrcontest}} comprises a task of word recognition for segmented historical documents, see  \cite{htrts2014} for all further details. 
These data consist of page images taken from the Bentham collection, a well-known transScriptorium project dataset.

Our neural networks have basically been used previously in the international handwriting competition OpenHaRT 2013 attached to the ICDAR 2013 conference, see \cite{citlab2013openhart}.
Moreover, with a system very similar to the one presented here, the CITlab team also took part in ICFHR's ANWRESH-2014 competition on historical data tables, see \cite{citlab2014anwresh} for the according system description.

Affiliated with the Institute of Mathematics at the University of Rostock, CITlab\footnote{\url{http://www.citlab.uni-rostock.de}} hosts joint projects of the Mathematical Optimization Group and PLANET intelligent systems GmbH, a small/medium enterprise focusing on computational intelligence technology and applications.
The work presented here was part of a common text recognition project 2010 -- 2014 and is extensively based upon PLANET's ARGUS software modules.

\section{Preprocessing}
\label{s:Preprocessing}
Firstly, we extract the line polygon sub-images as given in the XML-files provided for the HTRtS evaluation.
Those images are then subject to certain image preprocessing steps:
We start with a contrast normalization based on foreground/background pixel intensity levels.
In order to work properly, for our networks, it turned out to be particularly important to present all images in the same predefined pixel height during training and application. Here, we accomplish separate normalizations of the bottom, central and top image portion which finally results in a fixed height of 64 pixels for all images.

From the writings themselves, we remove disturbing substructures that might come in from adjacent lines by applying connected component clustering strategies. Finally, the average slant of the writing is normalized, too.

Being a special kind of preprocessing, it should be mentioned that we also presented slightly disturbed writing images during the network training \ref{ss:Training}. Technically, this noise arose from minor variations of specific parameters of the above-mentioned normalizations.
As usual, this is mainly done in order to enhance the systems' generalization capabilities.

\section{System specifications}
The following part describes the general architecture of the entire workflow, and we explain details for the core neural network and the decoding procedure.
Note that our two systems, \texttt{CITlab-Re-1} and \texttt{CITlab-Re-2,}
coincide in all structural respects. The only differences arise from their training and will be explained in subsection \ref{ss:Training}.

\subsection{Input}
\label{ss:Input}
The networks use entire writings as being prepared in the preprocessing described above in Section \ref{s:Preprocessing}.
In particular, there is no further segmentation.
As developed in \cite{graves2008offline}, every writing image is processed by reading its pixel data in four column-first ``directions'' that arise from combining top-down and bottom-up column traversals with left-to-right and right-to-left row traversals.

\subsection{Neural Network}
The neural networks used for CITlab systems are essentially based on preceeding work presented in \cite{graves2008offline}.
The basics are the same for all network types applied for HTRtS.
While preparing this contest, we investigated a variety of different modifications, parameter settings and certain random initializations, but mainly due to time limitations, this has not been a systematic search. Nevertheless, according to their specific performance on the HTRtS validation data set, we have chosen the two networks for the submissions mentioned above.

\subsubsection{Architecture}
\label{sss:Architecture}
The architecture of the neural nets particularly follows \cite{graves2008offline}, but we introduced essential modifications:
Instead of traditional MDLSTM cells, we use two layers with MDLeaky cells (\cite{citlab2013openhart}) which were shown to be more stable and thus yield better performance.
They embrace one layer of classical $\tanh$ cells in between.

Furthermore, this network core is preceeded by a first layer accomplishing \textsc{Gabor}-like feature extraction of frequency information with pre-defined, fixed parameters: 2 frequencies in 4 directions. In order to ensure realistic training durations, for HTRtS we did not use trainable weights in this bottom network part.

Finally, right before the output layer, there is a purely technical layer without trainable weights: It ensures that for every pixel column, just one activation value will be presented to the output neurons. Thus, the overall procedure results in a size reduction (subsampling) from the standard image height (see Section \ref{s:Preprocessing}) downto 1, but in fact, this is done step-wisely over the layers by consecutive shrinkings of factors 4 in y-direction and 3 or 2 in x-direction, respectively.

Altogether, the networks finally used in the HTRtS competition incorporate a total of 1.195 cells and 376.426 trainable weights.

\subsubsection{Output \& Training Algorithm}
\label{sss:Output}
According to the task of reading free texts, the networks have to deal with the complete alphabet of latin letters and digits along with usual special characters as punctuation marks, braces, quotes, \dots
In our setup, network output neurons are bijectively related to characters (or character classes in few exceptional cases), i.e. for the HTRtS task, we altogether work with 85 output neurons including an artificial garbage neuron that may e.g. indicate inter-character states.

The network output activation then should estimate the probability or confidence of the respective character at a certain image position.
Collecting those activations over time, i.e. over the entire writing image, finally yields the so-called \textsl{confidence matrix} as the final network output.

In order to let output activations really be useful confidence estimates, networks are trained by Backpropagation-Through-Time (BPTT) using the Connectionist Temporal Classification (CTC) algorithm described in \cite{graves2006connectionist} for calculating the gradient.
The initial internal weight values for those gradient descent training procedures are chosen randomly.

\subsection{Decoding}
\label{ss:Decoding}
After applying the standard softmax normalization, at each time step, the neural net provides a vector of probability estimates, each component counting for one entry of the alphabet. As it was mentioned before, collecting those vectors over time finally 
yields the network output, the so-called \textsl{confidence matrix}, ${N}(x)$, for a given input writing, $x$.
Decoding algorithms then typically search for a most likely word $w^*$ for the network output under consideration: 
\[w^*= \argmax_{w}p(w|N(x)) \,.\]
Since the garbage class typically has a high probability compared to other classes, the garbage ''letter'' is often cheap to insert which might bias the decoding result from shorter to longer word guesses. In order to correct for this, we prefer short words by considering an additional penalty term proportional to the word length $|w|$.
\begin{equation}
 \label{eq:MinCost}
 w^* = \argmin_{w}\;\; -\ln p(w|N(x)) - \alpha|w| \,.
\end{equation}
Here $\alpha$ is a constant which has been chosen according to experimental experience.

For decoding structured text lines, the CITlab group developed rather advanced and fairly elaborated strategies, algorithms and tools. While these will be presented in detail in upcoming publications, we only give an explanatory survey here. Note that, all of that is carried out by a Dynamic Programming based optimal path search in the confidence matrix subject to dictionary restrictions.

\begin{enumerate}[(1)]
	\item\label{i:part} We start by guessing rather conservatively which parts of the line might be single words or special characters.
	\item\label{i:oneWord} The respective submatrices are then decoded against the dictionary where the result gets a first rating.
	\item\label{i:twoWords} Since the first line partitioning may be faulty, we get a second rating by decoding under the assumption that the line segment under consideration were in fact containing two words.
	\item\label{i:oov} The third rating of the very same line segment simply is the \textsl{bestpath}, i.e. the sequence of characters that receive the highest confidence per network output time step.
	\item Take the result with the best rating out of (\ref{i:oneWord}, \ref{i:twoWords}, \ref{i:oov}).
		\\ Note that the out-of-vocabulary case is handled by (\ref{i:oov}): We simply take the raw network output if no dictionary word(s) receive a better rating.
	\item Append the special characters (i.e. punctuation marks, quotes, braces, \dots) according to the partitioning found in (\ref{i:part}).
\end{enumerate}

It seems worth noticing that CITlab's work focuses on the pure recognition part of the entire process. Hence in fact, no further models at character, word or language level were incorporated.
Consequently, in HTRtS we used the exact transcriptions contained in the XML-files for training and validation.
On the other hand, our core system does not consider the particular handling of special characters which was required in the HTRtS rules, i.e. separating some punctutation marks, certain quotes and braces from the words they were attached to in the original handwriting. Therefore, in order to get comparable evaluation results, we added a standard regular expression substitution: This should ensure a translation of our original recognition result into a HTRtS compatible form.

\subsection{Training}
\label{ss:Training}
CITlab's contribution to HTRtS fits to the restricted contest scheme, i.e. we exclusively exploited data provided by the HTRtS organizers for this contest.
We used (resp. extracted) three datasets, possibly except few items that, for some reason or the other, were unusable:

\begin{table}[ht]
	\centering
		\begin{tabular}[t]{lrl}
			data set & $\sharp$\,items & contents \\ \hline \\[-.75em]
			\texttt{HTRtS-full}  & 10.613 & HTRtS Training \& Validation decks \\
			\texttt{HTRtS-train} & 9.198 & full HTRtS Training deck \\
			\texttt{HTRtS-short} & 1.895 & all lines of \texttt{HTRtS-train} with at most 30 characters \\
		\end{tabular}
	\caption{CITlab's training data sets (cf. \cite{htrts2014}, Table 1)}
	\label{tab:CITlabTrainDataSets}
\end{table}
 
Networks were trained with a fixed momentum of 0.9 throughout the entire procedure.
As mentioned before, the two systems submitted use networks which only differ in their training setup, see the following Table \ref{tab:tableTrain} for details. Here one \textsl{epoch} refers to one time presenting the complete list used in any shuffled order.
\begin{table}[ht]
  \centering
  \begin{tabular}[t]{ccc}
    $\sharp\;$epochs & data set & \multicolumn{1}{c}{learning rate} \\
		\hline\hline \\[-.75em]
		\multicolumn{3}{l}{\texttt{CITlab-Re-1}} \\
		\hline \\[-.75em]
		40 & \texttt{HTRtS-short} & 2e-3 \\
		92 & \texttt{HTRtS-train} & 2e-3 \\
		21 & \texttt{HTRtS-train} & 1e-3 \\
		22 & \texttt{HTRtS-train} & 5e-4 \\
		30 & \texttt{HTRtS-full}  & 1e-3 \\
		30 & \texttt{HTRtS-full}  & 5e-4 \\
		\hline\hline \\[-.75em]
		\multicolumn{3}{l}{\texttt{CITlab-Re-2}} \\
		\hline \\[-.75em]
		40 & \texttt{HTRtS-short} & 2e-3 \\
		32 & \texttt{HTRtS-full}  & 5e-3 \\
		16 & \texttt{HTRtS-full}  & 2e-3 \\
		16 & \texttt{HTRtS-full}  & 1e-3 \\
  \end{tabular}
  \caption{Different training setup for the two submitted systems}
  \label{tab:tableTrain}
\end{table}

The entire training procedure of these networks usually lasts several weeks. Due to the technical setup of the distributed computing system CITlab uses, we cannot present more precise estimates of the computing time because they would all lack the necessary reliability.

By observing the training process continuously, we are able to adapt its parameters depending on intermediate performance results on the HTRtS validation partition. This explains some strangely looking figures in Table \ref{tab:tableTrain}.
However, it seems worth noting that throughout all these procedures, no overfitting has been observed.

\section{Application}
After describing the developed systems and the training procedure we conclude by summarizing the official evaluation and presenting  extended results we obtained in further, subsequent own examinations.

For testing various systems, we again exclusively used data provided by the HTRtS organizers:

\begin{table}[ht]
	\centering
		\begin{tabular}[t]{lrl}
			data set & $\sharp$\,items & contents \\ \hline \\[-.75em]
			\texttt{HTRtS-test}  & 860 & HTRtS Test deck \\
			\texttt{HTRtS-val} & 9.415 & HTRtS Validation deck \\
		\end{tabular}
	\caption{CITlab's test data sets (cf. \cite{htrts2014}, Table 1)}
	\label{tab:CITlabTestDataSets}
\end{table}
 
\subsection{Official Evaluation}
On the previously unknown evaluation dataset \texttt{HTRtS-test}, the official HTRtS calculations [\cite{htrts2014}] yielded a word error rate $\WER$ of 14.6\;\% and a character/label error rate $\CER$ of 5.0\;\% for the best CITlab-Re system.

Using the HTRtS organizers' evaluation tools, we (re)calculated these and other rates. Note that the tiny differences seem to be due to a slightly different approach in handling spaces. 
In order to see the systems' generalization abilities, we also calculated the respective rates for \texttt{HTRtS-val}. But note that, while these data were also used to train our systems, the decoder for the additional experiment only based on the dictionary extracted from \texttt{HTRtS-train}, i.e. it had to deal equally with out-of-vocabulary problems.
The following Table \ref{tab:EvaluationResults} presents all results.

\begin{table}[h]
	\centering
		\begin{tabular}[t]{l@{\hspace{3em}}*{2}{D{.}{.}{4}}c*{2}{D{.}{.}{4}}}
			& \multicolumn{2}{c}{$\WER$} && \multicolumn{2}{c}{$\CER$}\\ \cline{2-3} \cline{5-6} \\[-.75em]
			System & \multicolumn{1}{r}{\texttt{HTRtS-test}} & \multicolumn{1}{r}{\texttt{HTRtS-val}} && \multicolumn{1}{r}{\texttt{HTRtS-test}} & \multicolumn{1}{r}{\texttt{HTRtS-val}} \\ \hline \\[-.75em]
			\texttt{\texttt{CITlab-Re-1}} & 15.05 & 12.75 && 5.44 & 4.34 \\
			\texttt{\texttt{CITlab-Re-2}} & 14.48 & 10.78 && 5.06 & 3.57 \\
		\end{tabular}
	\caption{Extended evaluation results}
	\label{tab:EvaluationResults}
\end{table}

It seems worth noting that these results were obtained without any final word or language model application. This might explain the slightly different behaviour of the CITlab system compared to the others submitted to HTRtS: According to the diagrams plotted in Figures 4 or 5 of \cite{htrts2014}, such language or word models should be able to further reduce the error rates at central positions within lines or words, respectively.


\subsection{Perspective}
As already mentioned in Section \ref{ss:Decoding}, CITlab's core focus and abilities are mainly related to the recognition part of the entire handwritten document recognition process.
Obviously, it is particularly interesting to combine that with more advanced modeling techniques on word and language level.
However, due to time and resource limitations, we were not able to accomplish this truly compelling task within HTRtS.
But we remain very excited about future possibilities of combining CITlab's techniques, e.g. into the complete working pipeline provided by the HTRtS organizers \cite{htrts2014} from PRHLT at UPV.

\subsection*{Acknowledgement}
We are really indebted to the HTRtS organizers from the PRHLT group at UPV for setting up this evaluation and contest as well as for providing all the data along with their working chain and the evaluation tools.
In particular, we would like to thank Joan Andreu Sánchez for his ongoing help in all details and his patience in handling CITlab's questions and problems while preparing this HTRtS submission.

The work presented in this paper was funded by research grant no. V220-630-08-TFMV-S/F-059 (Verbundvorhaben, Technologieförderung Land Mecklenburg-Vorpommern) in European Social / Regional Development Funds.

\bibliographystyle{alpha}
\bibliography{SystemDescription_arXiv}

\begin{thebibliography}{GFGS06}

\bibitem[GFGS06]{graves2006connectionist}
Alex Graves, Santiago Fern{\'a}ndez, Faustino Gomez, and J{\"u}rgen
  Schmidhuber.
\newblock Connectionist temporal classification: labelling unsegmented sequence
  data with recurrent neural networks.
\newblock In {\em Proceedings of the 23rd international conference on Machine
  learning}, pages 369--376. ACM, 2006.

\bibitem[GS08]{graves2008offline}
Alex Graves and J{\"u}rgen Schmidhuber.
\newblock Offline handwriting recognition with multidimensional recurrent
  neural networks.
\newblock In {\em NIPS}, pages 545--552, 2008.

\bibitem[LGSL14]{citlab2014anwresh}
Gundram Leifert, Tobias Grüning, Tobias Strauß, and Roger Labahn.
\newblock {CITlab} {ARGUS} for historical data tables: Description of
  {CITlab}'s system for the {ANWRESH-2014} {Word Recognition} task.
\newblock Technical Report 2014/1, Universität Rostock, April 2014.
\newblock Available:
  \url{http://ftp.math.uni-rostock.de/pub/preprint/2014/pre14_01.pdf}.

\bibitem[LLS13]{citlab2013openhart}
Gundram Leifert, Roger Labahn, and Tobias Strauß.
\newblock {CITlab} {ARGUS} for arabic handwriting: Description of {CITlab}'s
  system for the {OpenHaRT} 2013 {Document Image Recognition} task.
\newblock In {\em Proceedings of the NIST 2013 OpenHaRT Workshop [Online]},
  August 2013.
\newblock Available: \url{http://www.nist.gov/itl/iad/mig/hart2013_wrkshp.cfm}.

\bibitem[SRTV14]{htrts2014}
Joan~Andreu Sánchez, Verónica Romero, Alejandro~H. Toselli, and Enrique
  Vidal.
\newblock {ICFHR2014 Competition on Handwritten Text Recognition on
  tranScriptorium Datasets (HTRtS)}.
\newblock In {\em {Proceedings of the International Conference on Frontiers in
  Handwriting Recognition -- ICFHR 2014}}, August 2014.

\end{thebibliography}

\end{document}